\documentclass[letterpaper, 10 pt, conference]{ieeeconf}  %

\IEEEoverridecommandlockouts                              %

\usepackage{cite}
\usepackage{graphicx}
\usepackage{caption}
\usepackage{subcaption}
\let\labelindent\relax
\usepackage[inline]{enumitem}
\usepackage{amsmath}
\usepackage{hyperref}
\usepackage{cleveref}
\usepackage{float}
\usepackage{amssymb}

\title{\LARGE \bf
CERBERUS: Simple and Effective All-In-One Automotive Perception 
Model with Multi Task Learning
}

\author{Carmelo Scribano$^{1,2,*}$, Giorgia Franchini$^{1}$, 
Ignacio Sañudo Olmedo,$^{3}$ and Marko Bertogna$^{1,3}$%
\thanks{$^{1}$Carmelo Scribano, Giorgia Franchini and Marko Bertogna are part of the Department of Physics, Informatics and Mathematics of the University of Modena and ReggioEmilia, 41125, Modena, Italy.
        {\tt\small \{name\}.\{surname\}@unimore.it}}%
\thanks{$^{2}$ Carmelo Scribano is part of the Department of Mathematical, Physical and Computer Sciences of the University of Parma, 43124, Parma, Italy.}%
\thanks{$^{3}$Ignacio Sañudo Olmedo and Marko Bertogna are part of HIPERT s.r.l, 41122, Modena, Italy.
        {\tt\small \{name\}\{surname\}@hipert.it}}%
\thanks{*Corresponding Author}%
}

\begin{document}

\maketitle
\thispagestyle{empty}
\pagestyle{empty}

\begin{abstract}

Perceiving the surrounding environment is essential for enabling autonomous or assisted driving functionalities. Common tasks in this domain include detecting road users, as well as determining lane boundaries and classifying driving conditions. Over the last few years, a large variety of powerful Deep Learning models have been proposed to address individual tasks of camera-based automotive perception with astonishing performances. However, the limited capabilities of in-vehicle embedded computing platforms cannot cope with the computational effort required to run a heavy model for each individual task.
In this work, we present CERBERUS (CEnteR Based End-to-end peRception Using a Single model), a lightweight model that leverages a multitask-learning approach to enable the execution of multiple perception tasks at the cost of a single inference.
The code will be made publicly available at \url{https://github.com/cscribano/CERBERUS}.

\end{abstract}

\section{INTRODUCTION}

Multi Task Learning  \cite{vandenhende2021multi} is a branch of machine learning in which several learning tasks are solved at the same time, while exploiting common and differences between the tasks. Developing lightweight yet powerful \textbf{Multi Task} models that can solve several perception tasks in a single forward pass, while maximizing the parameter sharing among the individual tasks, is an enabling capability to bring deep-learning based perception to production vehicles with limited computing resources. The recent introduction of the freely available dataset Berkeley Deep Drive (BDD100K) \cite{yu2020bdd100k} represents an important resource for the cause of multi-task perception models. Different Multi Task models \cite{wu2021yolop, vu2022hybridnets} have been lately proposed, showing encouraging results. \\
\noindent In this short paper we propose a Multi Task model that can simultaneously address the tasks of 1) road object detection (also classifying the object's occlusion), 2) lane estimation and 3) image classification for weather (sunny, rainy, cloudy etc..), driving scene (highway, city streets etc..) and day time (morning, night etc...).
Taking inspiration from modern single stage object detectors \cite{zhou2019objects, duan2019centernet, law2018cornernet}, we decided to cast both the task of object detection and lane estimation as regression of heatmaps (to encode the likelihood of the presence of objects or lanes at any spatial location) and offsets (used to decode the individual bounding boxes or lane instances).

The advantages of this election are manifolds: \begin{enumerate*}
  \item Employing the same representation among different tasks make the training process simple, allowing to even optimize for the same objective function.
  \item The anchor free approach allow for simple and efficient decoding of both the detection bounding boxes and the road marking lanes.
  \item The heatmap representation can be trivially extended to produce instance level prediction in a bottom-up fashion, as we demonstrate by incorporating the occlusion classification task in the object detection head.
\end{enumerate*} We design our model's architecture based on simple and well understood patterns, while focusing on a modular approach in order for the final model to be tailored accordingly to the deployment requirements. Leveraging a novel objective function for the heatmaps regression task, our model can be trained end-to-end while simultaneously optimizing for all the objectives. Finally, we tuned the model to reduce to the greatest extent possible the computational footprint, experimenting with efficient backbones \cite{tan2019efficientnet, sandler2018mobilenetv2}, and to ensure ease of deployment by refraining from leveraging exotic layers. 
The experiments presented suggest that the proposed approach represents a strong baseline and an important step forward towards a universal model for end to end perception.

\begin{figure}[!t]
    \centering
    \includegraphics[width=0.5\textwidth]{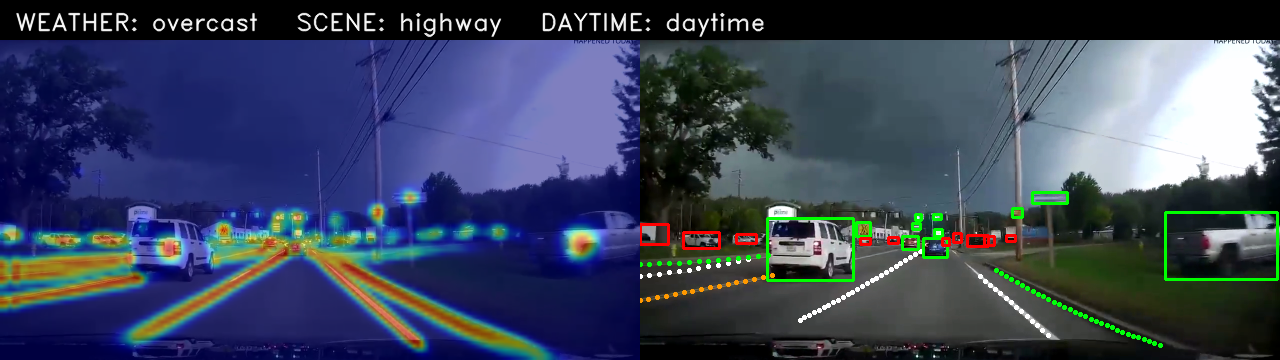}
    \caption{Qualitative inference results of CERBERUS on publicly available footage. (left) superimposed heatmaps (right) output for object detection and lane estimation.}
    \label{fig:demo}
\end{figure}
\section{RELATED WORKS}

\subsection{Object Detection}
Deep learning based object detectors can be classified in Multi-Stage and Single-Stage. The first kind of methods \cite{girshick2014rich, girshick2015fast, ren2015faster, cai2019cascade} include a first module to propose a large set of regions of the image that are likely to contain an object, the later stages then classify and refine the predictions. These methods have being known for being highly accurate, but computationally expensive. In opposition, single-stage detectors do not rely on a region proposal mechanism. In this sense, anchor-base detectors \cite{liu2016ssd, redmon2018yolov3, lin2017focal} output a set of \textit{candidate} boxes at predefined locations in the form of offset with respect to a
predefined set of anchor boxes, requiring an expensive process of non-maxima suppression to obtain the final detections. On the other hand, more recent anchor-free detectors \cite{zhou2019objects, duan2019centernet, law2018cornernet,zhou2019bottomup, tian2019fcos, carion2020end} output predictions only at object locations and thus require little post-processing. Those methods are often based on detection of keypoints, centerpoints or even on transformer decoders.

\subsection{Lane Estimation}
Lane estimation methods are divided in segmentation-based and detection-based: the first category \cite{neven2018towards,pan2018spatial} cast the lane estimation as a pixel-wise classification problem (either the pixel belongs to a lane or not), with the addition of a mechanism to associate the lane masks to separate marking instances. Detection-based methods instead have a lot in common with object detectors, as such some methods regress the lane instance using an anchor mechanism \cite{chen2019pointlanenet, xu2020curvelane}. Finally, recent works cast the lane estimation as a regression of keypoints and embeddings \cite{ko2021key} or offset \cite{wang2022keypoint} to aggregate the keypoints in distinct lanes.

\subsection{Multi-Task Approaches}
To the best of our knowledge, the closest works to our are the very recent YoloP \cite{wu2021yolop} and HybridNets \cite{vu2022hybridnets}. In our understanding, a main limitation of such models is the author's choice of relying on \textit{heterogeneous} representations for the individual tasks (e.g, combining a Yolo \cite{redmon2018yolov3} style object detection head with a segmentation head for the lane estimation). Such a choice makes the model inherently convoluted and the optimization phase harder because it relies on completely different objective functions.
\section{Method}\label{sec:method}

\subsection{Problem Formulation} 
As mentioned in the introduction, the proposed model is able to address multiple distinct tasks required for vision-based perception in a single inference step. In particular, given as input a single RGB image $I\in\mathbb{R}^{(w\times h\times 3)}$ we address the tasks of: \begin{enumerate}[label={\alph*)}]
\item Road Object Detection: predicting a bounding box and an associated class label for each object of among 10 distinct object classes. Additionally, for each object, we provide a binary label which indicates whether the object is fully visible or partially occluded.  
\item Lane Markings Estimation: regressing a polynomial curve and an associated class label for each lane marking visible. 
\item Image Tagging: providing three distinct multi-class classification labels associated to the whole frame, weather (7 classes), scene (7 classes) and time of day (4 classes).
\end{enumerate}
\noindent We train and evaluate our model on BDD100K. This publicly available dataset consists of 70K train images and 10K validation images sampled at 10Hz from a large set of 100K driving videos (around 40s each). Videos are captured in a wide range of scene types and conditions, allowing to train robust models able to generalize to real driving conditions. %

\subsection{Object detection}\label{subsec:object_detection}
\begin{figure}[H]
     \centering
    \includegraphics[width=0.5\textwidth]{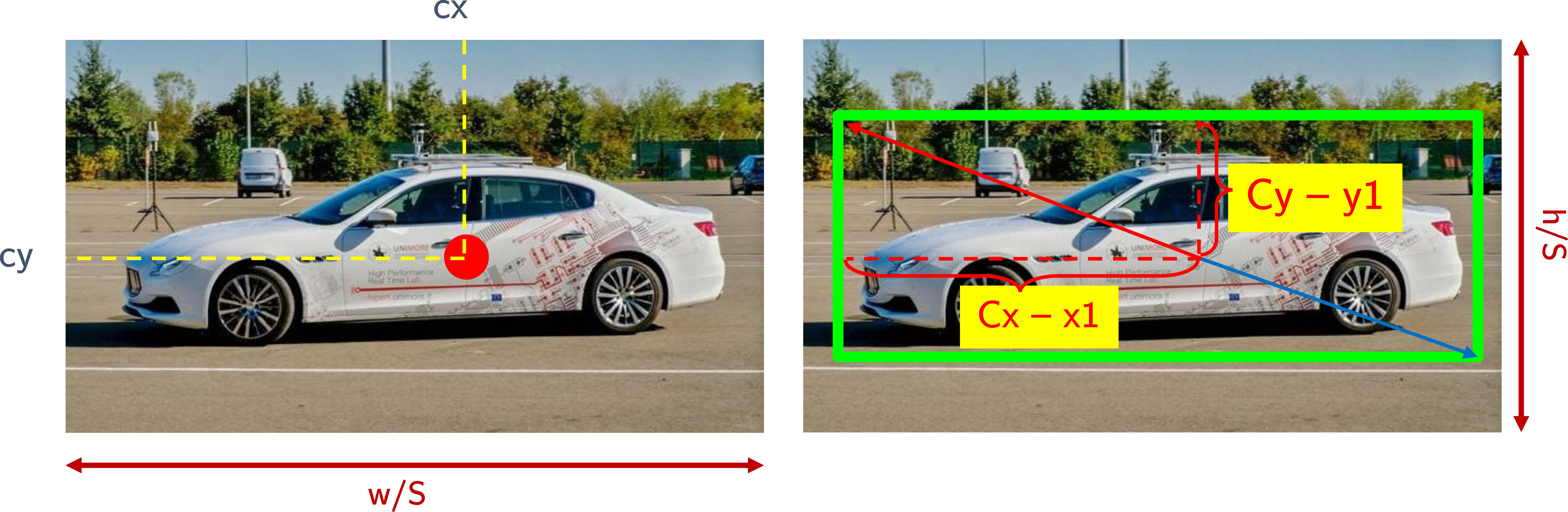}
    \caption{Definition of the object detection's targets: (left) object center-point (right) box-corners offsets: top-left (red) and bottom-right (blue).}
    \label{fig:objectdet}
\end{figure}

The desired output for the object detection task is a set of $N_O$ detections in the form $(x_1,y_1,x_2,y_2,c,o)$ where $x_1,y_1$ (resp. $x_2,y_2$) define the image space coordinate of the top-left (resp. bottom-right) corner of the bounding box, $c$ characterizes the object class and $o$ is the binary label for the object's occlusion state. Our approach is heavily based on the anchor-free detector described in \cite{zhou2019objects} which is based on keypoint estimation.\\
\noindent %
For each object class $k \in C_O$ with $|C_O|=10$,
we output a heatmap $\overline{H}_D^k \in \left(\frac{w}{S}\times\frac{h}{S}\right)$ (with $S$ being output stride) that encodes the center points of all the objects in $I$ belonging to class $k$. Given a set of detection ground truths, first the target keypoints in output space $c^i = \left(\frac{x_2^i-x_1^i}{2S}, \frac{y_2^i-y_1^i}{2S}\right) \textrm{ for } i \in \{0, \dots,N^k_O\}$ are computed as the geometrical center point of each box and rescaled with the output stride (rounding to the nearest integer), as in \cref{fig:objectdet} (left). Then the target heatmap  $H_D^k$  for the $k$-th object class is obtained by computing for each keypoint a Gaussian centered at the keypoint's location and taking the element-wise maximum:

\begin{equation}
\small
    \begin{gathered}
         H_D^k(x,y) = \max_j \left\{ \exp\left(-\frac{(x-c_x^j)^2+(y-c_y^j)^2}{\sigma^2}\right)\right\} \\
        \textrm{ for } j \in \{0,\dots ,N_O^k\}    \end{gathered}
    \label{eq:gt_heatmaps}
\end{equation}
with $\sigma$ being a hyperparameter.\\
In addition, we regress a double offset map (shared across all the object classes) $\overline{O}_D \in \left(w/S\times h/S\times 4\right)$. The values at the coordinates of each object center $c=(c_x, c_y)$ correspond to the box-corners offset vectors from the object center to the top-left and bottom-right corners of the target bounding box, as in \cref{fig:objectdet} (right). 
\begin{equation*}
    O_D(c_x,c_y) = (c_x-x_1/S, c_y-y_1/S,c_x-x_2/S, c_y-y_2/S)
\end{equation*}
\noindent We also regress the occlusion state for each detected object: this is simply achieved by predicting an occlusion map $V_D = \left(\frac{w}{S}\times\frac{h}{S}\times1\right)$ where the value at a center point's coordinates $(c_x,c_y)$ is meant to be 0 if the object is fully visible and 1 otherwise.

At inference time the predicted object detections for each class $k$ are retrieved by first taking the local maxima (the peaks of the Gaussians) from the keypoint heatmaps $\overline{P}^k = [(\overline{c}_x^1, \overline{c}_y^1)^k,...,(\overline{c}_x^{\overline{N}_O^k}, \overline{c}_y^{\overline{N}_O^k})^k]$, then the bounding boxes are reconstructed by summing the predicted box-corners offset vector for each center keypoint taken from $\overline{O}_D$ at the center point's coordinates. The occlusion classification is retrieved from $\overline{V}_D$ in the same way. %

\subsection{Lane Estimation} \label{subsec: lane_estimation}
\begin{figure}[ht]
     \centering
    \includegraphics[width=0.4\textwidth]{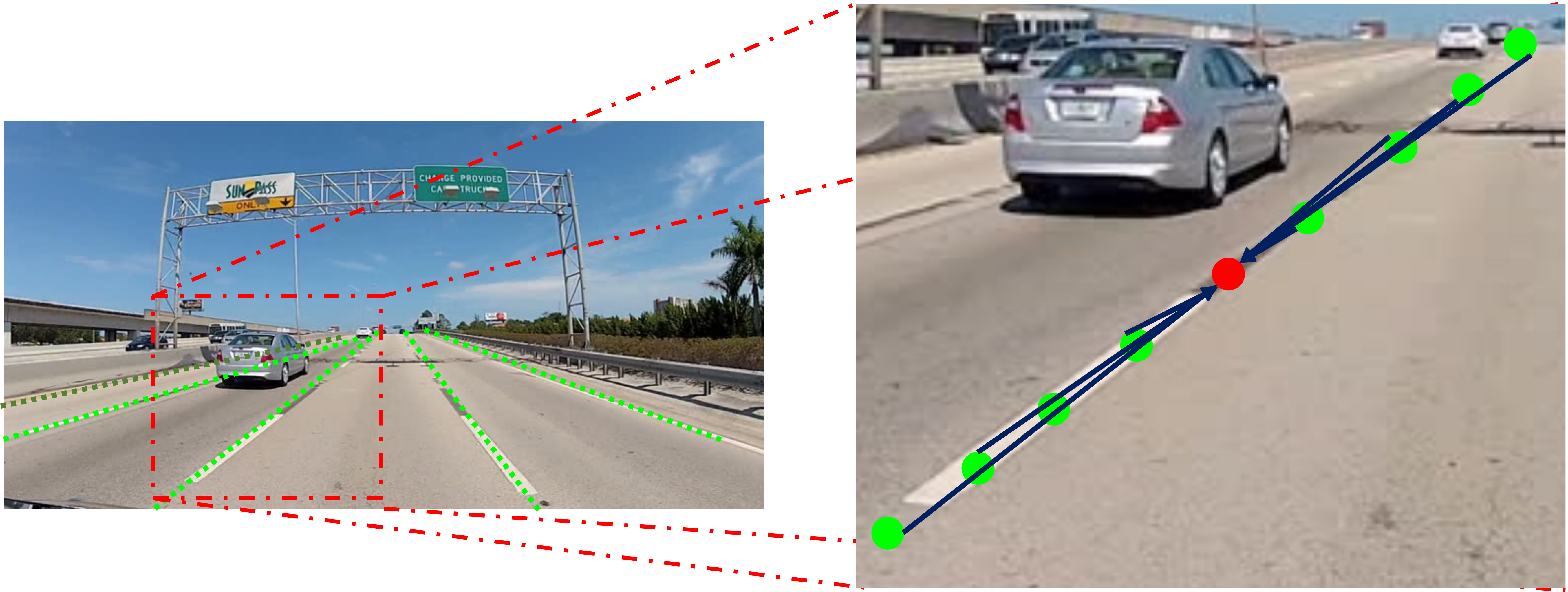}
    \caption{Representation of the lane estimation head used: each keypoint belonging to a line cast a vote offset for the mid-point of the lane that it belongs to.}
    \label{fig:lanest}
\end{figure}

The output of the lane estimation task is a set of lane instances for each lane class $l$ among the $N_l=8$ possible ones. The generic $i$-th lane instance $L_i^l = [(x_1, y_1),...,(x_{n^l_i},y_{n^l_i})]$ is a list of keypoint coordinates of arbitrary length $n_i^l$ (different for each lane instance). All the keypoints in $L_i^l$ should belong to a single lane and can be easily used to fit a polynomial curve representing the lane line boundaries in image space coordinates. As for the object detection we employ a keypoint-based approach, in particular our method is derived from \cite{wang2022keypoint} and \cite{ko2021key}. As shown in \cref{fig:lanest} this method allow to simultaneously recover all the keypoints belonging to any lane visible for the $l$-th lane class and regressing an offset vector for each keypoint that is used to effectively cluster the keypoints in distinct lane instances.

Like for the object detection we produce a heatmap $H_L^l$ for each lane class $l$, the target heatmap is obtained by first rescaling the keypoints coordinates to the output space and then fitting a Gaussian at each lane keypoint as from Eq. \ref{eq:gt_heatmaps}, disregarding of the lane instance. Additionally, we produce a single offset map $O_L \in (\frac{w}{S} \times \frac{h}{S}\times 2)$ where at each $j$-th keypoint location $p^i_j = (x^i_j, y^i_j)$ of the $i$-th lane instance we regress an  offset vector to the mid-point $L_i[|L_i| / 2]=(xm_i, ym_i)$ of the lane instance (\cref{eq:lane_ofs}).
\begin{equation}
    O_L(x^i_j, y^i_j) = (xm_i-x^i_j, ym_i-y^i_j)
    \label{eq:lane_ofs}
\end{equation}

At inference time the candidate lane keypoints $\left[(x_1,y_1),...,(x_{N_l},y_{N_l})\right]$ are first retrieved, for each lane class $l$, by selecting the Gaussian peaks from the predicted heatmap $\overline{H}_L^l$. Then for each candidate keypoint the corresponding lane center point $(\overline{xm}_i, \overline{ym}_i)_l$ is computed. Finally, the predicted lane midpoints are used to cluster the keypoints in individual lane instances using an agglomerative clustering algorithm \cite{ward1963hierarchical} with a fixed distance threshold.

\subsection{Image Tagging}
The image tagging task simply consists of three multi-class classification problems: the model is tasked to produce three labels $(C_w, C_s, C_t)$ corresponding to the frame-level label for weather, scene, and time of day respectively. %

\section{Model Architecture} 
\begin{figure}[!ht]
    \centering
    \includegraphics[width=0.5\textwidth]{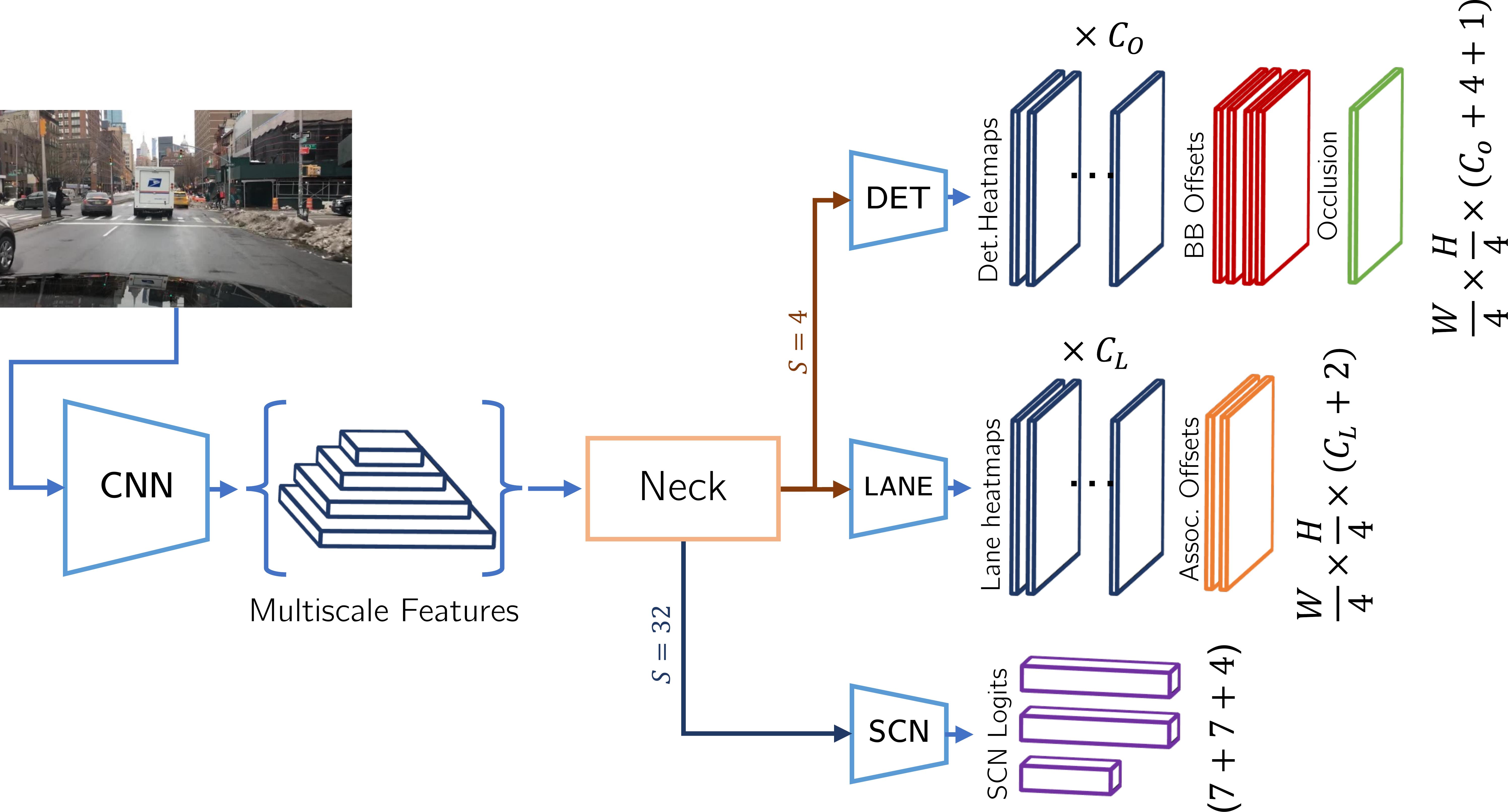}
    \caption{Overview of the proposed model. The outputs are produced by three distinct heads for object detection (DET), lane estimation (LANE) and scene classification (SCN).}
    \label{fig:architecture}
\end{figure}

Our model, depicted in \cref{fig:architecture}, follows a simple and robust approach common to several single-stage detector: first the convolutional part of an off-the-shelf classification model is used to extract a set of $n_F$ multiscale feature maps:
\begin{equation*}
    \mathcal{F}_i \in (C_i, w/S_i, h/S_i) \textrm{ for } i \in \{0,\dots ,n_F\}
\end{equation*}

With $C_i$ denoting the number of channels (increasing) and $S_i$ denoting the output stride (also increasing). For all of our models, we always select $n_F = 4$ output feature maps at fixed output strides $(4, 8, 16, 32)$, while the number of output channels $C_i$ is specific to the chosen backbone.\\
\noindent Subsequently, a \textit{neck} block further process the feature maps, increasing the output resolution and (optionally) aggregating the multiscale features. The neck provide as output a small resolution feature map ($S=32$), to be used exclusively for the image tagging task and a high resolution feature map ($S=4$) to be used for both the object and lane detection tasks. Finally, a distinct $head$ for each individual task apply a final convolutional layer in order to produce the task-specific output.

\subsection{Backbone}
In order to evaluate the flexibility of our approach we experimented with three different model families as convolutional backbones: \begin{enumerate*}
    \item Resnet \cite{he2016deep}, with its simple and well understood architecture, represents the most conservative option \item Mobilenet \cite{sandler2018mobilenetv2} instead represent a well established example of convolutional model with reduced memory and computational footprint \item finally EfficientNet \cite{tan2019efficientnet} is a more recent example of a model with low inference cost,  leveraging modern solutions. %
\end{enumerate*}\\
\noindent We purposely avoided to tinker with backbones either too computationally intensive to be effective in our intended deployment target or built with uncommon layers and architectural choices \cite{yu2018deep} that would make the conversion to the inference framework harder and potentially inaccurate. All the tested backbones are pretrained on the ImageNet classification dataset \cite{russakovsky2015imagenet}.

\subsection{Necks}\label{subsec:neck}
To process the feature maps obtained by the convolutional backbone, we experiment with two different neck blocks.\\

\noindent The \textbf{simple} neck only takes the last feature map (stride 32) and gradually increase the spatial resolution up to a stride of 4, required for object and lane detection, with alternating layers of regular 2D Convolution and Transposed convolution with stride of 2 and kernel size of 4 (implying a 2-time upsampling factor for each layer). The featuremap returned for the image tagging task is the output from the first convolutional layer. Furthermore, all the Transposed convolutional layers are initialized as bilinear interpolation kernels.

\begin{figure}[H]
    \centering
    \includegraphics[width=0.4\textwidth]{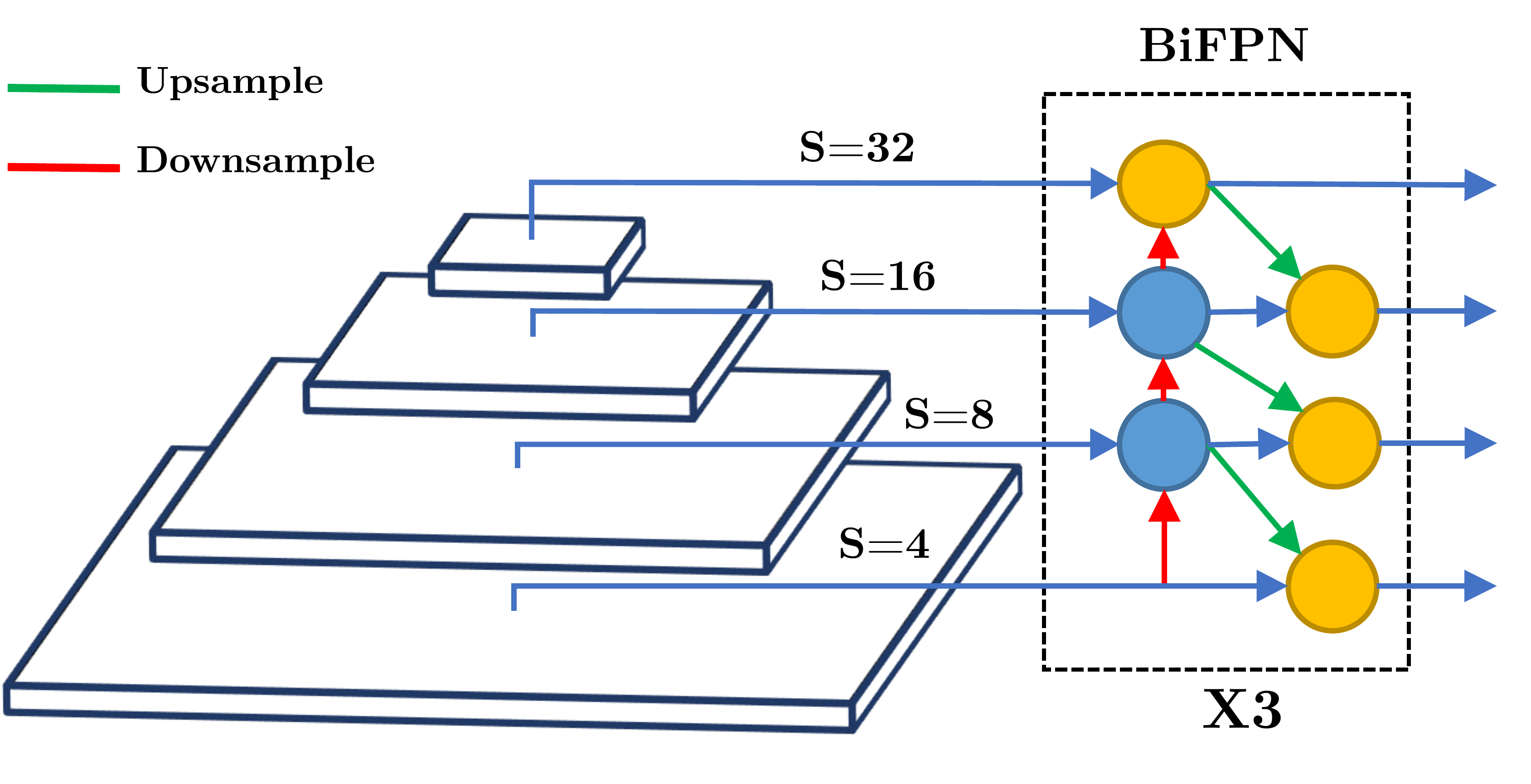}
    \caption{Overview of the BiFPN based head. In all the experiments, all the up sample operations are implemented as nearest neighbor interpolation and the down sample operations as MaxPooling layers.}
    \label{fig:bifpn}
\end{figure}

\noindent The more sophisticated, yet lighter, neck implementation is based on the Bi-directional Feature Pyramid Network (\textbf{BiFPN}) introduced in \cite{tan2020efficientdet}. This block is an evolution of the Feature Pyramid Layer (Fpn, \cite{lin2017feature}), it allows fusing feature maps ad different scales leveraging both top-down (up sampling) and a bottop-up (down sampling) paths to allow the information to flow in both directions, as depicted in \cref{fig:bifpn}. Given a set of multiscale feature maps $\mathcal{F}_S$ with $S$ denoting the stride, $(\mathcal{F}_4, \mathcal{F}_8, \mathcal{F}_{16}, \mathcal{F}_{32})$ the intermediate output for the top-down path is:
\begin{equation*}
    \mathcal{T}_S = 
        \begin{cases}
      \mathcal{F}_S & \text{if $S = \min_{S}$}\\
      Conv_{2D}(\mathcal{F}_S + up(\mathcal{F}_{S/2}) * w_\mathcal{T}^S) & \text{otherwise}\\
    \end{cases} 
\end{equation*}

\noindent Then the bottop-up path is defined as follows:

\begin{equation*}
    \mathcal{B}_S = 
        \begin{cases}
      \mathcal{T}_S & \text{if }S = \max_S\\
      Conv_{2D}(\mathcal{T}_S + down(\mathcal{T}_{2S}) * w_\mathcal{B}^S) & \text{otherwise}\\
    \end{cases} 
\end{equation*}

\noindent The trained weights $w_{\{\mathcal{B}|\mathcal{T}\}}^S$ are added to the feature fusion to allow the network to model the importance of each feature. We take $\mathcal{B}_4$ as input for the object and lane detection tasks, and $\mathcal{B}_{32}$ for the task of image tagging, respectively.

\subsection{Heads}
The heads follow a minimalistic design aimed at keeping the complexity overhead of each individual task to a minimum. The head for the lane estimation and object detection head share also the same input features and an identical structure: each of the desired outputs (i.e, heatmaps, offsets etc…) is obtained by applying a dedicated $3\times3$ convolutional layer to the feature map, followed by a ReLU activation, followed by a final $1\times1$ convolution to produce the desired number of output channels. A sigmoid activation is applied only to the heatmap's output layer, for training stability we found important to initialize the bias value of the last convolution to $4.6$ ($\sigma(4.6)=0.99$).\\
\noindent The image tagging head is as simple as it gets, a single $3\times3$ convolution is applied to the smallest feature map produced by the neck block, followed by activation and max-pooling, finally three fully-connected layer are applied to obtain the image level classification outputs.

\section{EVALUATION}

\begin{table*}[htbp]
\centering
\begin{tabular}{c|cc|cc|c|ccc}
\textbf{Model} & \textbf{Backbone} & \textbf{Neck} & \multicolumn{2}{c|}{\textbf{Object Detection}} & \textbf{Lane Estimation (IoU)} & \multicolumn{3}{c}{\textbf{Scene Classification (F1)}} \\
 &  &  & mAP50 (Boxes) & Accuracy (Occl) &  & Weather & Scene & Time of Day \\ \hline\hline
RN34-sim                        & Resnet34                           & Simple                         & 0.210            & 0.913              & 0.640                                           & 0.815         & 0.790      & 0.932            \\
RN34-bifpn                      & Resnet34                           & BiFPN                          & 0.265            & 0.910              & 0.644                                           & 0.817         & 0.791      & 0.932            \\
RN50-bifpn                      & Resnet50                           & BiFPN                          & 0.274            & 0.911              & 0.648                                           & 0.822         & 0.794      & 0.935            \\
RN101-bifpn                     & Resnet101                          & BiFPN                          & 0.281            & 0.912              & 0.65                                            & 0.825         & 0.794      & 0.934            \\
EnB2-bifpn                      & Efficientnet b2                    & BiFPN                          & 0.272            & 0.907              & 0.647                                           & 0.821         & 0.790      & 0.937            \\
Mobbv2-bifpn                    & MobilenetV2                        & BiFPN                          & 0.230            & 0.9                & 0.633                                           & 0.813         & 0.790      & 0.934           
\end{tabular}
  \caption{Results of various configurations for all the tasks evaluated on BDD100K}
  \label{tab:results}
\end{table*}

\subsection{Experimental Setup}
The annotation provided by BDD100K for the object detection task already comes in the desired format, where each bounding box is represented in image space as the coordinates of  top-left and bottom-right corners plus an object label class. Hence the heatmap and offset targets can be easily computed as detailed in \cref{subsec:object_detection}, with the value of $\sigma$ computed as in \cite{law2018cornernet}. Other competitors (\cite{wu2021yolop, vu2022hybridnets}) select only specific classes and squash them in a single "vehicle`` super class. Instead, we preserve the class as provided by the dataset.
\noindent For the lane estimation task the provided lane markings are annotated by labeling the two edges of each lane line as two distinct lines, while for our method we would rather predict a single line in the middle of each road marking. We are first required to associate and merge the two road marking annotation in a single one, then we can obtain a keypoint array for each lane instance by sampling the Bézier curves with a fixed sampling pace, finally the targets can be computed as in \cref{subsec: lane_estimation}, with $\sigma$ fixed to $2$\\

\noindent To keep the computational demands low, the RGB images, with an original resolution of $1280\times720$ pixels, are first downscaled with a factor of 2 and then cropped to a final resolution of $640\times320$. The output stride $S = 4$ for all the experiments. During training several data augmentation strategies are adopted in order to improve generalization against changes in appearance and illumination (random noise, RGB shift, random brightness and contrast…) and camera positioning (random affine transforms, random crop, random horizontal flips). %

\subsection{Objective Functions}
Multitask deep learning models are often non-trivial to train due to instability and some tasks dominating the gradient of the loss. Hence, careful weighting of the individual tasks and often elaborate multi-stage training strategies are commonly required. Instead, our model can be trained end-to-end mostly relying on basic objective function, except for the loss function used for the heatmaps regression. In total, we use three different loss functions:\\

\subsubsection{Heatmap Regression Loss}
It is the same for the two tasks, commonly used objective function for heatmaps regression include the standard $L1$ and MSE ($L2$) distance functions,  other loss function explicitly engineered to boost heatmap regression include the Wing Loss \cite{feng2018wing, feng2020rectified}. Lately the Focal Loss introduced in \cite{lin2017focal} has been found to be useful for this matter \cite{law2018cornernet, duan2019centernet, zhou2019objects}, the focal loss add a weighting term to the Cross-Entropy Loss (usually used in binary classification problems)  to avoid the contribution from the large amount easy to classify values (i.e, background pixels) to dominate the gradient magnitude. In our setup, we experience severe instability issue with the Focal Loss (as well as with Generalized Focal Loss \cite{li2020generalized}). Driven by a similar principle we instead propose to use a weighted version of $L2$ distance reported in \cref{eq:hmloss}
\begin{equation}
    \footnotesize \mathcal{L}_H(H, \overline{H}) = \frac{1}{N_K}\sum max\{(1+H)^\alpha, (1+\overline{H})^\beta\}(H-\overline{H})^2
    \label{eq:hmloss}
\end{equation}
Recalling that both $H$ and $\overline{H}$ are tensors of shape $(w/S \times h/S)$, the elementwise weighting factor will give an exponentially larger relevance to locations where either the target heatmap or the predicted heatmap are closer to the maximum value of $1$. In this way the vast heatmap background (filled with zeros) has a low relevance, the target weight with exponential $\alpha$ allow focusing on hard values where a peak is present, the prediction weighting term with exponential $\beta$ is instead responsive of penalizing false positives. For all our experiments, we fix $\alpha=4$ and $\beta=2$.\\

\subsubsection{Offsets Regression Loss}
The objective function for the offsets required by the formulation of both tasks is computed only at the corresponding keypoint location $(kx,ky)$ as a simple $L1$ distance. Other works suggest that a more sophisticated function as IoU \cite{zhou2019iou} or DIoU loss \cite{zheng2020distance} could boost the performance of the object detection task, we did not further investigate these alternatives. While practically those objective functions could be leveraged in our formulation, we opted for the simplest loss function to avoid incurring in hard to track down convergence issues.\\

\subsubsection{Classification Loss}
Finally, we use the standard Cross-Entropy Loss function both for the image tagging task and for the (binary) object level occlusion classification. The predicted object occlusion logit is readout at the ground-truth boxes center points values from the occlusion map, $\overline{O}_D$ and the loss value is then computed for each object and averaged out.

\subsection{Experiments and Evaluation}
\begin{table}
\centering
\begin{tabular}{c|cc|cc}
\textbf{Model} & \textbf{Params (M)} & \textbf{MACCs (G)} & \multicolumn{2}{c}{\textbf{FP16 Infer (ms)}} \\
 &  &  & Xavier & Nano \\ \hline
RN34-sim & 29.1 & 61.67 & 14.34 & 205.64 \\
RN34-bifpn & 22.2 & 19.77 & 8.85 & 103.98 \\
RN50-bifpn & 24.6 & 21,8 & 11.09 & 130.06 \\
EnB2-bifpn & 8,7 & 7.45 & 14.11 & 135.30 \\
Mobbv2-bifpn & 2.7 & 5.9 & 5.86 & 68.37
\end{tabular}
  \caption{Inference results: inference times reported account only for the forward pass computation, without accounting for the decoding phase.}
  \label{tab:inference}
\end{table}

All the experiments reported in \cref{tab:results} are run with the same set of hyperparameters: in particular we train for 75 epochs with a batch size of 64 (for a total of over $160K$ steps) using the optimizer Adam, the learning rate is linearly increased from zero to $2.5^{-4}$ during $3500$ steps, then is halved after $100K$ steps. For comparison in \cref{tab:results} we report the results of six different architectures with different backbones: Resnet (34, 50 and 101), MobilenetV2 and Efficientnet-b2. For the neck (\cref{subsec:neck}) we report a single experiment using the simple version, due to the obvious superiority of the BiFPN Neck.\\

\subsubsection{Performance evalutation}
We evaluate the performance for the tasks of Object detection using the popular mAP@0.5 metric \cite{lin2014microsoft}, considering all the classes of the dataset, we extract from the predictions all the bounding boxes whose confidence score (value at the keypoint location) is greater or equal to $0.25$. The object occlusion sub-task is evaluated by first matching the ground truth bounding boxes to predicted ones, solving a minimum weight matching problem, then Accuracy metric is computed only for the successfully matched boxes.\\
\noindent To evaluate the lane estimation, we generate binary masks for both the true and the estimated lane instances (disregarding the class label). This choice introduces in the final error the noise added by the decoding and polynomial fitting process. In a future iteration we will likely be focusing on a different metric less dependent on the post-processing stage. Finally, for the image tagging task we individually evaluate the three multi-class classification tasks computing the $F1$ score across all classes.\\

\subsubsection{Profiling}
To evaluate the efficiency of our model we report in \cref{tab:inference} measures of number of parameters and the total number of multiply-accumulate operations. More important, we collect inference times on real deployment scenarios by running the models of two popular embedded platforms of the Nvidia Tegra family: the AGX Xavier represents the high-end segment, being capable of up to 32 TOPs with a power consumption of up to 20W. Meanwhile, the Jetson Nano represents the lowest end of the spectrum, with only 0.5 TOPs and 5W of power consumption. For both devices we replicate the same setup by running the last version (4.6.1 at the time of writing of the Jetpack SDK and executing the models in 16-bits floating point precision using the proprietary inference framework TensorRT (version 8.2.1).
\section{CONCLUSIONS}
Due to the preliminary nature of this short paper we cannot draw firm conclusions. It is utterly clear that more experiments and in particular detailed comparison with other methods will be required. However, the experimental results are enough to conclude that our methodology achieve the proposed goal of addressing all the considered perception tasks at a low inference cost, while exhibiting satisfactory performance and being easily and effectively deployable on embedded hardware.

\addtolength{\textheight}{-12cm}   %

\small
\section*{ACKNOWLEDGMENT}
This work has been partially supported by the project ``CEMP" funded by Region Lombardia, Italy (POR-FESR 2014-2020 - Call HUB Research and Innovation) and by the INdAM research group GNCS.%

\bibliographystyle{IEEEtran}
\bibliography{bib}
\end{document}